\documentclass[10pt,twocolumn,letterpaper]{article}

\usepackage{iccv}
\usepackage{times}
\usepackage{epsfig}
\usepackage{graphicx}
\usepackage{amsmath}
\usepackage{amssymb}
\usepackage{booktabs}   
\usepackage{arydshln}   
\usepackage{adjustbox}  
\usepackage{multirow}   
\usepackage{caption}

\usepackage[pagebackref=true,breaklinks=true,letterpaper=true,colorlinks,bookmarks=false]{hyperref}

\iccvfinalcopy 

\def\iccvPaperID{****} 

\ificcvfinal\fi

\usepackage[dvipsnames]{xcolor}

\usepackage{xcolor}
\usepackage[normalem]{ulem}

\newcommand\vcenterhead[1]
  {%
    \multirow{2}{*}[-0.5\dimexpr\aboverulesep+\belowrulesep+\cmidrulewidth\relax]{#1}%
  }

\newcommand{\nus}{nuScenes}
\newcommand{\name}{BEVFusion4D}

\begin{document}

\title{BEVFusion4D: Learning LiDAR-Camera Fusion Under Bird’s-Eye-View via Cross-Modality Guidance and Temporal Aggregation}

\author{Hongxiang Cai\thanks{Equal Contribution.}
\and
Zeyuan Zhang\footnotemark[1]
\and
Zhenyu Zhou\footnotemark[1]
\and
Ziyin Li\footnotemark[1]
\and
Wenbo Ding\thanks{Corresponding author.}
\and
Jiuhua Zhao\\
SAIC AI LAB\\
{\tt\small \{caihongxiang,zhangzeyuan,zhouzhenyu,liziyin,dingwenbo,zhaojiuhua\}@saicmotor.com}
}

\maketitle
\ificcvfinal\thispagestyle{empty}\fi

\graphicspath{{\subfix{../Images/}}}

\begin{abstract}

Integrating LiDAR and Camera information into Bird's-Eye-View (BEV) has become an essential topic for 3D object detection in autonomous driving. Existing methods mostly adopt an independent dual-branch framework to generate LiDAR and camera BEV, then perform an adaptive modality fusion. Since point clouds provide more accurate localization and geometry information, they could serve as a reliable spatial prior to acquiring relevant semantic information from the images. Therefore, we design a LiDAR-Guided View Transformer (LGVT) to effectively obtain the camera representation in BEV space and thus benefit the whole dual-branch fusion system. LGVT takes camera BEV as the primitive semantic query, repeatedly leveraging the spatial cue of LiDAR BEV for extracting image features across multiple camera views. Moreover, we extend our framework into the temporal domain with our proposed Temporal Deformable Alignment (TDA) module, which aims to aggregate BEV features from multiple historical frames. Including these two modules, our framework dubbed BEVFusion4D achieves state-of-the-art results in 3D object detection, with 72.0\% mAP and 73.5\% NDS on the nuScenes validation set, and 73.3\% mAP and 74.7\% NDS on nuScenes test set, respectively.
\end{abstract}

\graphicspath{{\subfix{../Images/}}}

\section{Introduction}


\begin{figure}[htbp]
\centering
\includegraphics[width=0.5\textwidth]{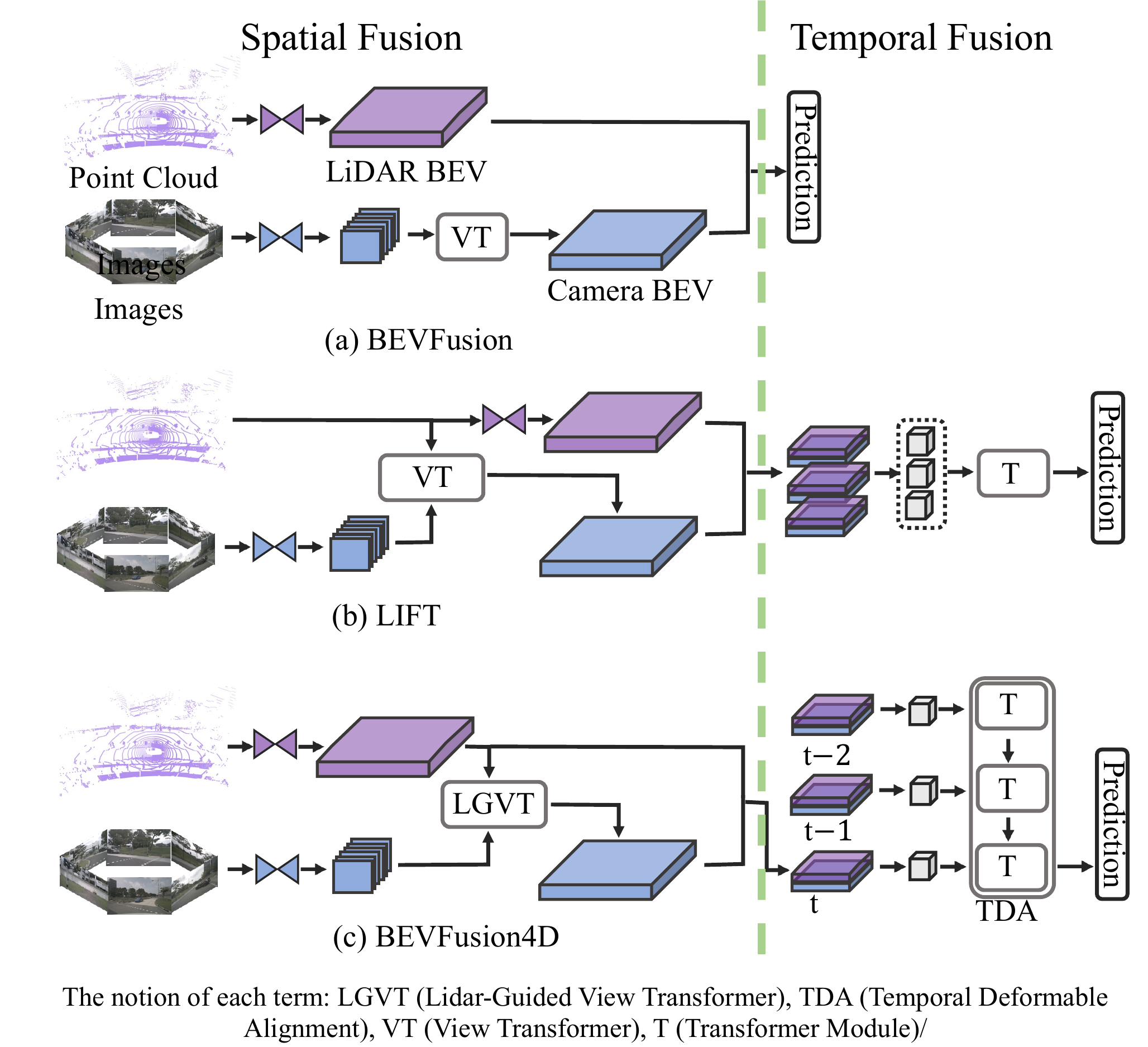}
\caption{Comparison of our BEVFusion4D with relevant works BEVFusion~\cite{liangbevfusion, liu2022bevfusion} and LIFT~\cite{zeng2022lift}. In BEVFusion (a), a dual-branch framework is introduced to process the information of point clouds and images separately and unify the BEV features in the spatial domain. LIFT (b) presents a spatiotemporal fusion paradigm that first generates BEV grids of LiDAR and camera features and then fuses them with a global self-attention module. In our method (c), we propose two modules LGVT and TDA to tackle spatiotemporal fusion sequentially. LGVT learns to generate camera BEV with the guidance of encoded LiDAR features. Furthermore, TDA recurrently involves fused spatial features from previous frames to aggregate spatiotemporal information. }
\label{Fig: teaser} 
\end{figure}

The task of 3D object detection has gathered enormous attention throughout the years and become commonplace when it comes to autonomous driving. As two fundamental modalities, LiDAR and camera sensors are capable of acquiring the surroundings in different manners. Point clouds generated by LiDAR sensors perceive the scene via the emitted wavelength. It excels in depicting the accurate spatial position of the object and offering reliable geometric information. Compared to point clouds, image data records a highly detailed picture of the scene and carries denser semantic and texture information. Therefore, it is recognized as a critical complement to the former sensor. 
Despite the challenge, incorporating sensing information from two distinct modalities is considered a research field of high value.


Due to the innate characteristics of LiDAR and camera, it remains an investigation to effectively integrate representations of the two modalities. Query-based methods such as TransFusion~\cite{bai2022transfusion} propose a two-stage pipeline to associate LiDAR and image features sequentially. However, the performance of the system depends largely on the query initialization strategy. Recent works~\cite{liangbevfusion, liu2022bevfusion, li2022unifying} have proved the effectiveness of a dual-branch fusion framework. Shown in Fig.~\ref{Fig: teaser}(a), the encoded feature from the backbone is transferred and unified in an intermediate feature representation such as Bird-Eye-View (BEV)~\cite{ma2022vision}, 
 although this paradigm has gained popularity in society, the difficulty of the camera in perceiving geometry information limits the impact of the camera branch and deters the system from fusing semantic information in the imagery data.

\begin{figure}[htb]
\centering
\includegraphics[scale=0.5]{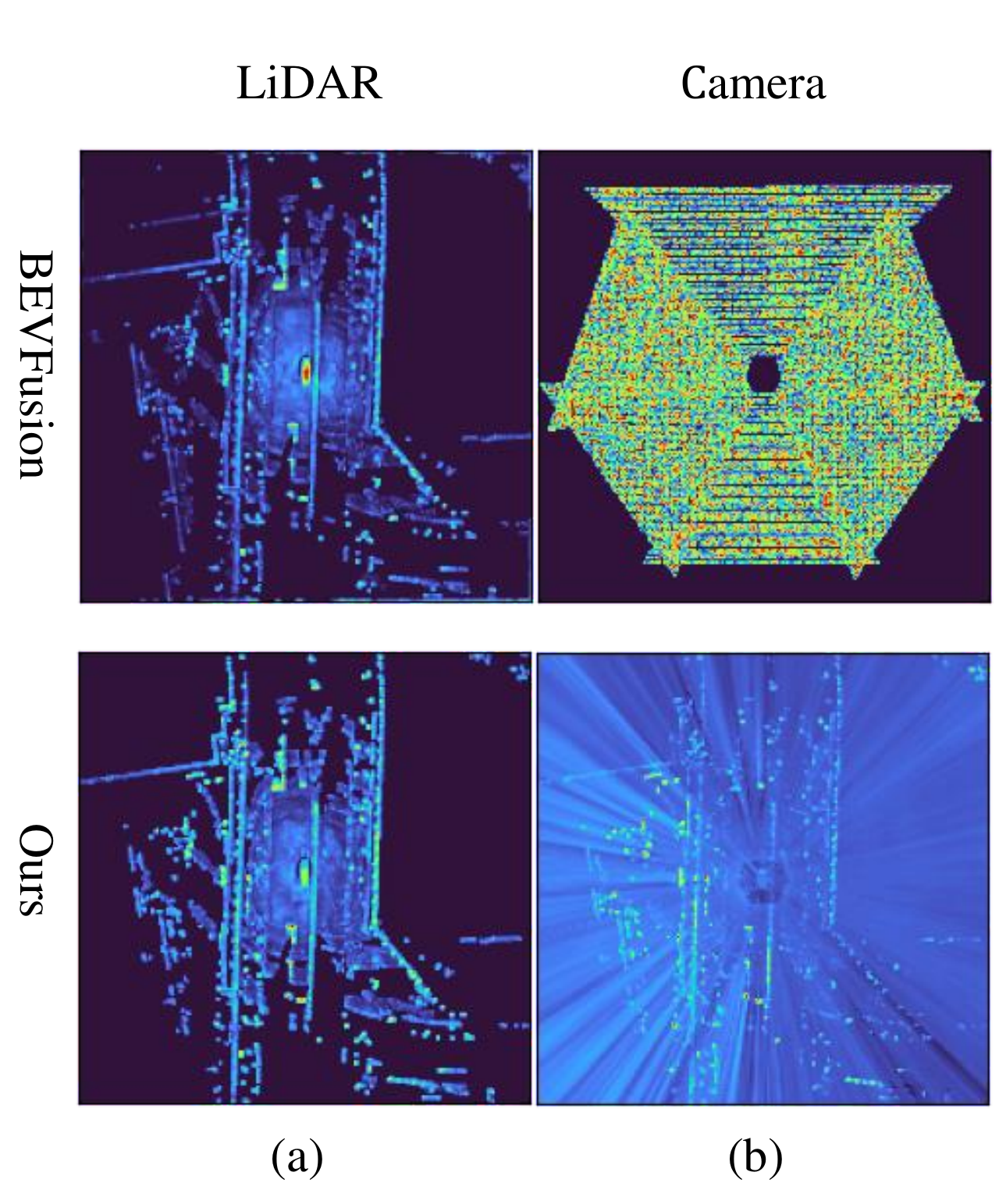}
\caption{Visualization of LiDAR and camera BEV features generated from BEVFusion~\cite{liangbevfusion} and ours. Our camera BEV feature contains more spatial information than BEVFusion, which demonstrates that the spatial prior of LiDAR plays a critical role in helping generate camera BEV feature.}
\label{Fig.bev_compare} 
\end{figure}

To make effective use of the camera data and further maintain semantic information during the fusion, we propose a simple yet effective solution that seeks to augment the camera BEV feature with explicit guidance of concurrent LiDAR. In the illustrated Fig.~\ref{Fig: teaser}(c), we design an attention-based camera view transformer named LGVT. It learns to effectively acquire the semantic information of the target conditioned on the prior of LiDAR. Since the point clouds of LiDAR describe a relatively accurate spatial distribution of the scene, they can be used as a critical prior to calibrating relevant semantics of the target, thereby facilitating more valuable information for the fusion. In Fig.~\ref{Fig.bev_compare}, we compare our visualization result of camera BEV with BEVFusion~\cite{liangbevfusion}. Benefiting from the spatial prior of LiDAR, the scene outline and target location could be easily distinguished in our camera BEV feature, which is invisible in BEVFusion.

On the other hand, the complementary views and motion cues encoded in the adjacent frames motivate recent works to integrate temporal information into the framework~\cite{li2022bevformer, huang2022bevdet4d}. As one pioneer, LIFT~\cite{zeng2022lift} marked the initial attempt to exploit the spatiotemporal information in the fusion framework. As shown in Fig.~\ref{Fig: teaser}(b), it takes the input 4D data as an integrated entity and directly aggregates the sequential cross-sensor data by the classical transformer architecture. However, the fusion system suffers from a computational overload due to the global self-attention mechanism. We instead fuse spatial and temporal features in two separate modules to significantly reduce the overall cost and lower the complexity of integrating spatiotemporal information. Moreover, to efficiently aggregate spatially-fused BEV features of consecutive frames, we propose a temporal fusion module based on the deformable attention strategy~\cite{zhu2020deformable}. With the capability of dynamic relationship modeling and sparse sampling, our proposed module learns to associate temporal features across long spans in a cost-effective way. We also demonstrate that our module offers the potential to align the temporal feature of moving objects without explicit motion calibration.

In this paper, we present BEVFusion4D together with two dedicated modules in spatial and temporal domains to facilitate multi-frame cross-modal information aggregations. We summarize our contribution into three folds:

\par 1) We propose BEVFusion4D, an efficient fusion framework to learn holistic feature representations in spatiotemporal domains. 
\par 2) Our proposed Lidar-Guided View Transformer (LGVT) aims to generate efficient camera BEV by a spatial prior, which facilitates the propagation of semantic information and benefits feature fusion in the spatial domain. Furthermore, we also design a temporal deformable alignment module (TDA) to cover supplementary information from the adjacent frames.
\par 3) We conducted extensive studies on the detection track of nuScenes datasets to validate the performance of our approach and observed a consistent lead in both spatial-only and spatiotemporal cases.

\graphicspath{{\subfix{../Images/}}}

\section{Related Works}
\subsection{3D Object Detection with Single Modality}
 Single modality-based 3D detection aims to predict the object information as a form of the 3D bounding box, given sensing results from only one main source such as LiDAR point clouds or camera images. Methods based on the LiDAR sensor input either regress object prediction directly from a scattered set of points~\cite{li2021lidar, qi2018frustum, yang20203dssd} or convert them into a unified grid. For example, Point-based approaches such as PointNet~\cite{qi2017pointnet} consume a point cloud and aggregate its feature from an end-to-end network. Another group of works approaches the problem from a perspective of pillar encoding~\cite{lang2019pointpillars, shi2022pillarnet}, voxel-encoding~\cite{zhou2018voxelnet} or range image-encoding~\cite{fan2021rangedet, sun2021rsn} and processes the encoded feature via a detector. In contrast, camera-based methods generally extract the view-dependent dense image features via a 2D image backbone and compose a unified 3D feature representation such as Bird-Eye-View (BEV) for 3D prediction. Derived from~\cite{philion2020lift, reading2021categorical}, encoded 2D image features are projected into a 3D frustum grid by predicting a discrete depth distribution for each feature point. BEVDepth~\cite{li2022bevdepth} further enrolls depth supervision to render a better feature frustum. Other recent works~\cite{wang2022detr3d, li2022bevformer, jiang2022polarformer, liu2022petr} enlightened by DETR~\cite{carion2020end} propose to model the interaction between image features via the transformer architecture and yield final predictions through a set of object queries. While most approaches gained substantial improvement in the field, aggregating information from multiple sensor sources is generally beyond their fields.

\subsection{3D Object Detection with Multi-Modalities}
The recent trend also casts a rising light on multi-modality-based 3D detection and believes that information from multiple sources will compensate each other and benefit the overall detection results. As one of the pioneers, PointPainting~\cite{vora2020pointpainting} opens up an approach of point-level fusion where the retrieved image semantic information from a semantic segmentation network is appended into the input points and thus realizes the modality interaction. This is further extended by various methods such as \cite{wang2021pointaugmenting, Yin2021MVP, huang2020epnet}. Different from the former, the work of~\cite{Yoo20203DCVFGJ, Xu2021FusionPaintingMF} represents a new approach to enable cross-modal relationships at the feature level. Recent works attempt to associate corresponding modality features from various concepts. BEVFusion~\cite{liangbevfusion,liu2022bevfusion} derives an effective framework that adopts modality fusion in BEV space. Methods like~\cite{bai2022transfusion, li2022unifying, chen2022autoalignv2} take the insight of DETR-like structure~\cite{carion2020end, zhu2020deformable}, leveraging attention mechanism to enforce an adaptive relationship between features.~\cite{yang2022deepinteraction} proposes a bilateral representation to progress information exchange. Compared to the existing approaches, our work inspired by the BEV-based solutions maintains the overall framework simplicity but instead achieves a competing performance through the dedicated query module and temporal fusion strategy.

\subsection{Temporal Fusion in 3D Object Detection}
Temporal information is valued as providing motion cues and complementary views of the object. This motivates various proposals to explore temporal fusion in the object detection task. For example, 4D-Net~\cite{Piergiovanni_2021_ICCV} develops a network with dynamic connection learning to gradually fuse temporal LiDAR and RGB image sequences. Huang~\textit{et al.}\cite{huang2020lstm} exploits the effect of an LSTM in modeling the temporal feature correlations. Moreover, the works of~\cite{qi2021offboard, yang20213d} apply a single-frame detector to extract object proposals and further combine proposal features from multiple frames. PETRV2~\cite{liu2022petrv2} extends the 3D position encoding to achieve temporal alignment and fuses image features in time. BEVFormer~\cite{li2022bevformer} manages to build the interaction of spatial and temporal image information through the popular deformable transformer. LIFT~\cite{zeng2022lift} proposes a spatial-temporal fusion framework for multi-modal sensor fusion based on the global self-attention mechanism. However, directly fusing 4D features using this strategy will bring an unnecessary computation cost and discourage efficient information exchange. In contrast, we achieve temporal integration on top of the spatially fused BEV features, which significantly lessens the computing budget. Inspired by BEVFormer~\cite{li2022bevformer}, we also adopt a deformable transformer to effectively incorporate features through a temporal sparse sampling strategy.


\graphicspath{{\subfix{../Images/}}}

 \begin{figure*}[htbp]
\centering
\includegraphics[width=\textwidth]{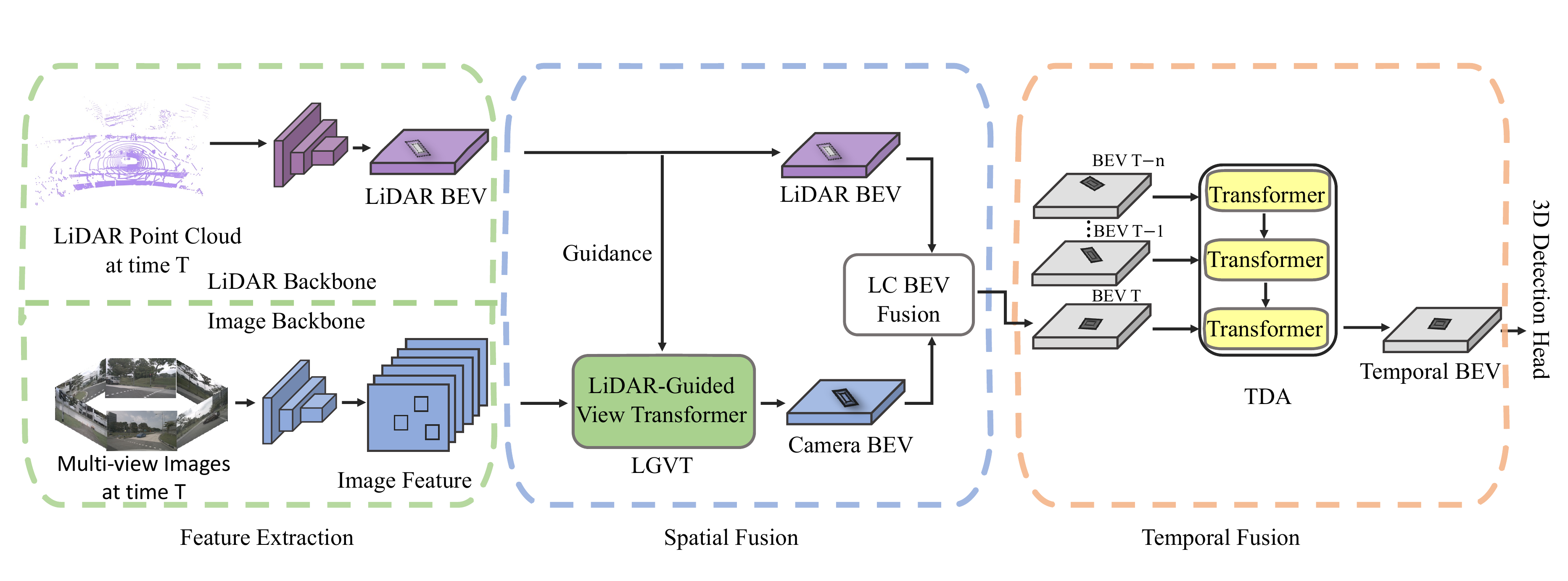}
\caption{The overall pipeline of our spatiotemporal fusion framework consists of feature extraction, spatial fusion and temporal fusion stages. The input of LiDAR Point Cloud and camera Images are processed via a separate feature extraction backbone. For spatial fusion, our proposed LiDAR-Guided View Transformer (LGVT) fuses multi-view image features into camera BEV conditioned on the spatial prior of LiDAR. Then spatial fusion is presented to unify multi-modality BEV features. For temporal fusion, the historical information of spatially-fused BEV features is aggregated through a temporal deformable alignment module (TDA) to achieve spatiotemporal feature interactions.}
\label{Fig.main} 
\end{figure*}

\section{Method}
We first give an introduction of feature extraction in Sec.~\ref{sec3.1}. Sec.~\ref{sec3.2} provides the procedures of spatial fusion with a detailed explanation of our LGVT module. The aforementioned two stages are conducted identically on each temporal frame. Furthermore, we describe our temporal fusion method based on the proposed TDA module (Sec.~\ref{sec3.3}).
\subsection{LiDAR and Camera Features Extraction} \label{sec3.1}
In the feature extraction stage, we adopt a dual-branch paradigm of previous works~\cite{liangbevfusion, liu2022bevfusion} to process cross-modality data. Specifically, the LiDAR point cloud and multi-view images at the current time are fed into two independent backbones to form high-level feature representations. For the LiDAR stream, the input point clouds $P \in \mathbb{R}^{N \times D}$ is converted into a unified grid through voxelization~\cite{zhou2018voxelnet} and further processed by the 3D sparse convolution~\cite{yan2018second} to form the feature representation in the BEV space $B_{LiDAR} \in \mathbb{R}^{X \times Y \times C}$, where $X$, $Y$, $C$ represent the size of the BEV grid and feature dimensions. The image backbone takes multi-view image data to yield 2D image feature $I \in \mathbb{R}^{N_{c} \times C\times H\times W}$, where $N_{c}$, $C$, $H$, $W$ denote the number of cameras, feature dimensions, image height, and image width, respectively. Then the fusion of extracted LiDAR and image features will be carried out in the following spatial fusion stage.

\subsection{Spatial BEV Features Fusion} \label{sec3.2}
The extracted features of LiDAR and images deliver the essential information of geometry and semantics, respectively. To further incorporate these features into a unified BEV space, a view transformation is required to project the multiple 2D image features into camera BEV space. Previous methods~\cite{liangbevfusion, liu2022bevfusion} choose LSS~\cite{philion2020lift} to achieve this task by lifting 2D features into 3D space with various depth probabilities. However, the absence of reliable depth supervision for this module usually results in an inferior performance. In contrast, we rely on the relatively accurate spatial information from the pre-trained LiDAR BEV to acquire corresponding semantic features. This enables our proposed LGVT to effectively project the 2D image features into BEV space.

\begin{figure*}[htb]
\centering
\includegraphics[width=1\textwidth,height=0.4\textwidth]{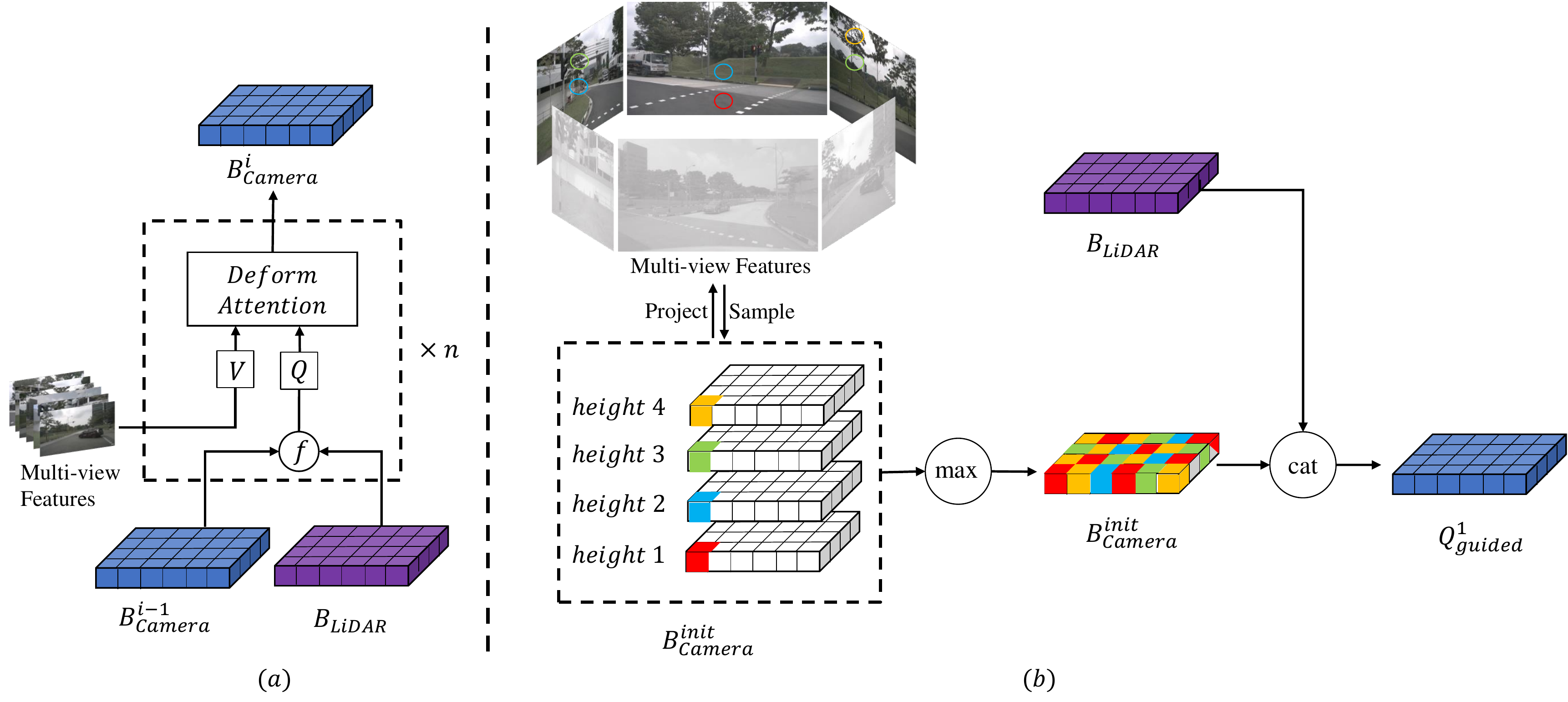}
\caption{(a) is the architecture of LGVT. (b) is the process of camera BEV features initialization, where different colors on $B^{init}_{Camera}$ represent pre-defined heights, and circles on multi-view features are the sampling location computed by Eq.~\ref{eq.project}.}
\label{Fig.img_BEV} 
\end{figure*}

\textbf{LiDAR-Guided View Transformer Module}. LGVT takes advantage of the deformable attention module to transform 2D image features into BEV features under the guidance of LiDAR spatial prior. Fig.~\ref{Fig.img_BEV} shows the LGVT module's architecture. In the $i$-$th$ layer, the module fuses the previous layer camera BEV feature $B_{Camera}^{i-1}$ of and LiDAR BEV features $B_{LiDAR}$ into query feature $Q_{guided}^i$, then performs deformable cross-attention~\cite{zhu2020deformable} with 2D image features $I$ to update camera BEV features, the LGVT module can be formulated as follows: 
\begin{equation} \label{eq.query_generation}
\begin{split}
    Q_{guided}^i = f(B_{LiDAR}, B_{Camera}^{i-1}), \\
    \text{where}~B_{Camera}^{i-1} = 
    \left\{\begin{matrix}
    B_{Camera}^{init},&i=1,\\
    B_{Camera}^{i-1} ,&i>1.
    \end{matrix}\right.
\end{split}
\end{equation}
\begin{equation}
    B_{Camera}^i = DeformCrossAttn(Q_{guided}^i, I),
\end{equation}
where $B_{Camera}^{i} \in \mathbb{R}^{X \times Y \times C}$ is the camera BEV feature in the $i_{th}$ layer, $f$ is concatenate operation.

\textbf{Camera BEV Feature Initialization}. At the first layer, we pre-define $N$ heights for an empty camera BEV feature and initialize it by filling each BEV grid with the corresponding 2D image features. Specifically, for an empty camera BEV feature grid $B_{Camera}^{init}{(i, j)}$ with coordinate $(i, j)$, we pre-define $N$ heights $\{h_1, h_2,...,h_N\}$ for this grid and project $\{(i, j, h_1),...(i, j, h_N)\}$ to 2D multi-view image features using intrinsics and extrinsics. Then we sample the maximum image feature of each height and compute the average of valid image features in all views. Finally, the camera BEV will be initialized with sampled image features. The process can be formulated as follows: 
\begin{gather}
    ((x_n^{1}, y_n^{1}),...,(x_n^{M}, y_n^{M})) = Project(i, j, h_n), \label{eq.project}\\
    I_{(x_n^m, y_n^m)} = Sample((x_n^m, y_n^m), I)),  \label{eq.sample}\\ 
    B_{Camera}^{init}{(i, j)} = \frac{1}{M_{hit}}\sum_{m=1}^{M_{hit}}{\max_{n=1,...,N}(I_{(x_n^m, y_n^m)})}, \label{eq.sample_selection}
\end{gather}
where $M$ is the number of views, $M_{hit}$ is the number of valid sampling location, $n=1,...,N, m=1,...,M$, $(x_n^m, y_n^m)$ are the coordinates in 2D image features corresponding to the $(i, j, h_n)$ at $m$-${th}$ view, $Project$ is the projection matrix and $Sample$ is the bilinear interpolation process to sample the 2D image feature at $(x, y)$.

\textbf{LiDAR-Camera BEV Features Fusion}. As common practice in~\cite{liangbevfusion, liu2022bevfusion}, we simply concatenate the LiDAR and camera BEV features and send the result to convolution layers. The fusion process can be formulated as follows:
\begin{equation} \label{eq:spatialFusion}
    F = Conv(Concat(B_{LiDAR},B_{Camera})).
\end{equation}

\subsection{Temporal BEV Features Fusion} \label{sec3.3}
\begin{figure}[htbp]
\centering
\includegraphics[scale=0.35]{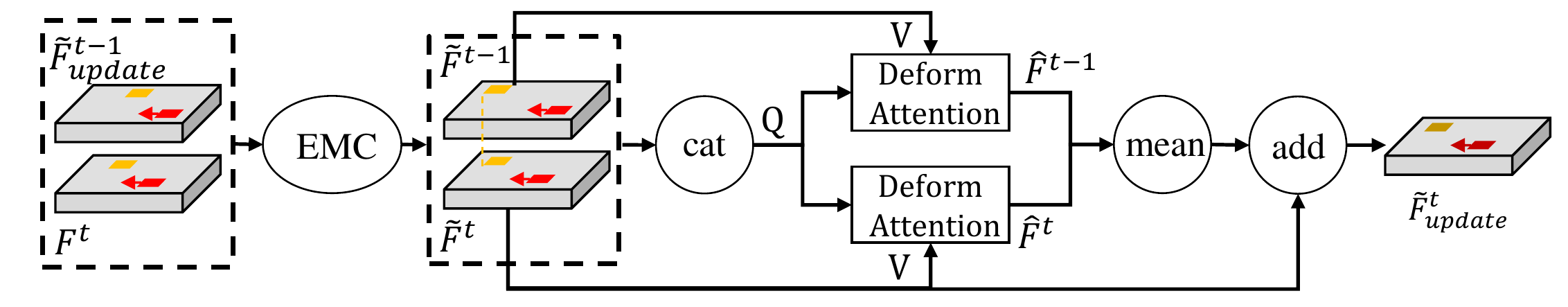}
\caption{Temporal BEV features fusion module. The red and orange squares represent moving and stationary objects, respectively, while the arrows indicate the direction of the object's movement. The EMC module is the Ego Motion Calibration formulated by Eq.~\ref{eq.alignment}.}
\label{Fig.temporal_bev} 
\end{figure}

The target's historical location and orientation information is beneficial for its current motion estimation. Besides, temporal information can also help detect remote, near or occluded objects in current time. Thus BEVFusion4D employ temporal fusion on the historical BEV features. In detail, in a recurrent way, previous BEV feature is first calibrated with the current one according to ego-motion information, then moving objects between two frames are further aligned in the temporal deformable alignment module to obtain the temporal fused BEV feature.

\textbf{Ego Motion Calibration}. Fig.~\ref{Fig.temporal_bev} shows the architecture of our temporal BEV features fusion module. Fused BEV features stored in chronological order are denoted as $\{F^{T-n},..., F^{T}\}$.  BEVFusion4D aligns each BEV feature $F^{t}$ with the next BEV features $F^{t+1}$ based on the motion information of the ego vehicle in temporal. This calibration process can be described by the following formula:
\begin{equation} \label{eq.alignment}
    \Tilde{F}^t = K^{t \to {t+1}}~F^t, t={T-n},...,{T-1}.
\end{equation}

The matrix $K^{t \to {t+1}}$ are termed as ego-motion transformation matrix aligned with $F^{t+1}$ at time $t$.

The calibration process described by Eq.~\ref{eq.alignment} is limited to aligning stationary objects due to the lack of motion information. Consequently, moving objects may exhibit motion smear without any calibration, and as historical BEV features increase, the motion smear effect can become more pronounced. 

\textbf{Temporal Deformable Alignment}. To alleviate this challenge, we recurrently adopt the deformable attention mechanism to establish the correspondence between consecutive frames. The deformable attention mechanism is designed to adaptively learn the receptive field, thus enabling it to effectively capture the salient features of moving objects in the aligned BEV features. By exploiting the deformable attention mechanism, we can significantly reduce the motion smear and enhance the accuracy of the alignment process. We propose the Temporal Deformable Alignment(TDA) to conduct temporal fusion specific to moving objects.

We denote $\Tilde{F}, \hat{F}$ as BEV features calibrated by Eq.~\ref{eq.alignment} and temporal fused by TDA, respectively. Firstly, TDA concatenate two consecutive calibrated frames $\Tilde{F}^{t-1}_{update}$ and $\Tilde{F}^{t}$ as $\Tilde{F}^{t-1,t}$. Then the deformable attention mechanism is applied to obtain $\hat{F}^{t-1}$ and $\hat{F}^{t}$ by using $\Tilde{F}^{t-1,t}$ as the query and $\Tilde{F}^{t-1}$, $\Tilde{F}^{t}$ as values. Next, TDA computes the average of $\hat{F}^{t-1}$ and $\hat{F}^{t}$ element-wise to update frame $\Tilde{F}^t$ by adding it to frame $\Tilde{F}^t$. The updated frame $\Tilde{F}^t_{update}$ can then be used for the fusion of the subsequent BEV feature. The process can be formulated as follows:
\begin{gather}
    \Tilde{F}^{t-1,t} = Concat(\Tilde{F}^{t-1}_{update}, \Tilde{F}^{t}), \\
    \hat{F}^{t-1} = DeformAttn(\Tilde{F}^{t-1,t},\Tilde{F}^{t-1}), \\
    \hat{F}^{t} = DeformAttn(\Tilde{F}^{t-1,t},\Tilde{F}^{t}), \\
    \Tilde{F^t}_{update} = \Tilde{F^t} + average(\hat{F}^{t-1}, \hat{F}^{t}),
\end{gather}
where $t=2,...,T$.

\graphicspath{{\subfix{../Images/}}}

\begin{table*}[t]
\setlength{\tabcolsep}{2.6pt}
	\centering
    \captionsetup{width=.82\textwidth}
	\caption[STH]{\textbf{Evaluation Results on the \nus{} validation set.} \name{}-S indicates our framework without the temporal fusion module. The blue and red colors represent the second-best results and best results, respectively. 
 	}
    \vspace{0.1cm}
    \label{tab:nuscenes_val}
	\begin{tabular}{l|c|cc|cccccccccc}
    \hline
         \small{Method} & \small{Data} & \small{mAP$\uparrow$} & \small{NDS$\uparrow$} & \scriptsize{Car} & \scriptsize{Truck} & \scriptsize{C.V.} & \scriptsize{Bus} & \scriptsize{Trailer} & \scriptsize{Barrier} & \scriptsize{Motor.} & \scriptsize{Bike} & \scriptsize{Ped.} & \scriptsize{T.C.} \\
    \hline
        {FUTR3D}~\cite{chen2022futr3d} &LC &64.2 &68.0  &86.3  &61.5  &26.0  &71.9  &42.1  & 64.4 & 73.6 & 63.3 &82.6 &70.1 \\
        {AutoAlignV2}~\cite{chen2022autoalignv2} &LC &67.1 &71.2 & - & - & - & - & - & - & - & - & - & - \\        
        {BEVFusion}~\cite{liu2022bevfusion} &LC &68.5 &71.4  &-  &-  &-  &-  &-  &- &- &- &- &- \\           
        {BEVFusion}~\cite{liangbevfusion} &LC &69.6 &72.1  &89.1  &66.7  &30.9  &77.7  &42.6  & \textcolor{blue}{\textbf{73.5}} &79.0 &67.5 &89.4 &79.3 \\        {DeepInteraction}~\cite{yang2022deepinteraction} &LC &69.9 &72.6  & 88.5 &64.4  &30.1  &79.2  &44.6  & \textcolor{red}{\textbf{76.4}} &79.0 &67.8 &88.9 &80.0 \\ 
        \hdashline
        \name{}-S &LC & \textcolor{blue}{\textbf{70.9}} & \textcolor{blue}{\textbf{72.9}} & \textcolor{blue}{\textbf{89.8}} & \textcolor{blue}{\textbf{69.5}} & \textcolor{blue}{\textbf{32.6}} & \textcolor{blue}{\textbf{80.6}}	& \textcolor{blue}{\textbf{46.3}} & 71.0 & \textcolor{blue}{\textbf{79.6}} & \textcolor{blue}{\textbf{70.3}} & \textcolor{blue}{\textbf{89.5}} & \textcolor{blue}{\textbf{80.3}} \\
        \name{} & LCT & \textcolor{red}{\textbf{72.0}} & \textcolor{red}{\textbf{73.5}} & \textcolor{red}{\textbf{90.6}} & \textcolor{red}{\textbf{70.3}} & \textcolor{red}{\textbf{32.9}} & \textcolor{red}{\textbf{81.5}} & \textcolor{red}{\textbf{47.1}} & 71.6 & \textcolor{red}{\textbf{81.5}} & \textcolor{red}{\textbf{73.0}} & \textcolor{red}{\textbf{90.2}} & \textcolor{red}{\textbf{80.9}} \\ \hline              
        \multicolumn{14}{l}{\scriptsize{The notion of 
        class: Construction vehicle (C.V.), pedestrian (Ped.), traffic cone (T.C.). The notion of Data: Camera (C), LiDAR (L), Temporal (T).}}\\        
    \end{tabular}
\end{table*}

\begin{table*}[t]
\setlength{\tabcolsep}{2.6pt}
	\centering
    \captionsetup{width=.82\textwidth}
	\caption[STH]{\textbf{Evaluation Results on the \nus{} test set.} \name{}-S indicates our framework without the temporal fusion module. The blue and red colors represent the second-best results and best results, respectively. 
 	}
    \vspace{0.1cm}
    \label{tab:nuscenes_test}
	\begin{tabular}{l|c|cc|cccccccccc}
    \hline
         \small{Method} & \small{Data} & \small{mAP$\uparrow$} & \small{NDS$\uparrow$} & \scriptsize{Car} & \scriptsize{Truck} & \scriptsize{C.V.} & \scriptsize{Bus} & \scriptsize{Trailer} & \scriptsize{Barrier} & \scriptsize{Motor.} & \scriptsize{Bike} & \scriptsize{Ped.} & \scriptsize{T.C.} \\
    \hline
        {PointPillars~\cite{Lang2019PointPillarsFE}} & L  & 30.5 & 45.3 & 68.4 & 23.0 & 4.1 & 28.2 & 23.4 & 38.9 & 27.4 & 1.1 & 59.7 & 30.8 \\
          {CBGS~\cite{Zhu2019ClassbalancedGA}} & L  & 52.8 & 63.3 & 81.1 & 48.5 & 10.5 & 54.9 & 42.9 & 65.7 & 51.5 & 22.3 & 80.1 & 70.9 \\
          {CenterPoint~\cite{yin2021center}} & L  & 60.3 & 67.3 & 85.2 & 53.5 & 20.0 & 63.6 & 56.0 & 71.1 & 59.5 & 30.7 & 84.6 & 78.4 \\
         {TransFusion-L~\cite{bai2022transfusion}} & L  & {65.5} & {70.2} & {86.2} & {56.7} & {28.2} & {66.3} & {58.8} & {78.2} & {68.3} & {44.2} & {86.1} & {82.0} \\
         \hdashline
          {PointPainting~\cite{vora2020pointpainting}} & LC   & 46.4 & 58.1 & 77.9 & 35.8 & 15.8 & 36.2 & 37.3 & 60.2 & 41.5 & 24.1 & 73.3 & 62.4 \\
          {3D-CVF~\cite{Yoo20203DCVFGJ}} & LC   & 52.7 & 62.3 & 83.0 & 45.0 & 15.9 & 48.8 & 49.6 & 65.9 & 51.2 & 30.4 & 74.2 & 62.9 \\
          {PointAugmenting~\cite{wang2021pointaugmenting}} & LC   & 66.8 & 71.0 & {87.5} & 57.3 & 28.0 & 65.2 & 60.7 & 72.6 & 74.3 & 50.9 & 87.9 & 83.6 \\
          {MVP~\cite{Yin2021MVP}} & LC  & 66.4 & 70.5  & 86.8 & 58.5 & 26.1 & 67.4 & 57.3 & 74.8 & 70.0 & 49.3 & {89.1} & 85.0 \\
          {FusionPainting~\cite{Xu2021FusionPaintingMF}} & LC   & 68.1 & 71.6 & 87.1 & {60.8} & 30.0 & {68.5} & {61.7} & 71.8 & {74.7} & {53.5} & 88.3 & 85.0 \\
        {TransFusion~\cite{bai2022transfusion}} & LC  & {68.9} & {71.7} & 87.1 & 60.0 & {33.1} & 68.3 & 60.8 & {78.1} & 73.6 & 52.9 & {88.4} & 86.7 \\ 
       {AutoAlignV2~\cite{chen2022autoalignv2}} & LC  & 68.4 & 72.4 & 87.0 & 59.0 & 33.1 & 69.3 & 59.3 & - & 72.9 & 52.1 & 87.6 & - \\   
        {BEVFusion}~\cite{liu2022bevfusion} &LC &70.2 &72.9  &-  &-  &-  &-  &-  &- &- &- &- &- \\        
        {BEVFusion~\cite{liangbevfusion}} & LC  & {71.3} & {73.3} & 88.1 & 60.9 & 34.4 & 69.3 & 62.1 & 78.2 & 72.2 & 52.2 & 89.2 & 86.7 \\          
        {DeepInteraction~\cite{yang2022deepinteraction}} & LC  & 70.8 & 73.4 & 87.9 & 60.2 & 37.5 & 70.8 & 63.8 & \textcolor{blue}{\textbf{80.4}} & 75.4 & 54.5 & \textcolor{red}{\textbf{91.7}} & \textcolor{blue}{\textbf{87.2}} \\ 
         {\name{}-S} &LC & \textcolor{blue}{\textbf{71.9}} & \textcolor{blue}{\textbf{73.7}} & \textcolor{blue}{\textbf{88.8}} & \textcolor{blue}{\textbf{64.0}} & \textcolor{blue}{\textbf{38.0}} & \textcolor{blue}{\textbf{72.8}} & \textcolor{blue}{\textbf{65.0}} & 79.8 & \textcolor{blue}{\textbf{77.0}} & \textcolor{blue}{\textbf{56.4}} & 90.4 & 87.1\\ 
         \hdashline 
         {LIFT~\cite{zeng2022lift} } &LCT &65.1 &70.2 &87.7 &55.1 &29.4 &62.4 &59.3 &69.3 &70.8 &47.7 &86.1 &83.2 \\ 
         {\name{} } & LCT & \textcolor{red}{\textbf{73.3}} & \textcolor{red}{\textbf{74.7}} & \textcolor{red}{\textbf{89.7}} & \textcolor{red}{\textbf{65.6}} & \textcolor{red}{\textbf{41.1}} & \textcolor{red}{\textbf{72.9}} & \textcolor{red}{\textbf{66.0}} & \textcolor{red}{\textbf{81.0}} & \textcolor{red}{\textbf{79.5}} & \textcolor{red}{\textbf{58.6}} & \textcolor{blue}{\textbf{90.9}} & \textcolor{red}{\textbf{87.7}} \\ 
        \hline
        \multicolumn{14}{l}{\scriptsize{The notion of 
        class: Construction vehicle (C.V.), pedestrian (Ped.), traffic cone (T.C.). The notion of Data: Camera (C), LiDAR (L), Temporal (T). }}\\
	\end{tabular}
\end{table*}

\section{Experiments}
In this section, we present experimental results to reason the principle of our framework. On top of this, several ablation studies are displayed to illustrate the benefits of our two model components.
\subsection{Experimental Setups}
\noindent\textbf{Datasets and metrics.}
We conduct various experiments and evaluate our methods on the nuScenes dataset. The nuScenes dataset is a large-scale multimodal dataset benchmarked for the autonomous driving study \cite{caesar2020nuscenes}. In the task of 3D Objection, nearly 1000 scenes are provided with a division of 700 training, 150 validation, and 150 testings, respectively. For each scene, a segment of 20s duration is recorded by 6 cameras and 1 LiDAR. We conduct our experiments on the annotated keyframes of each sensor. The calibrated cameras captured 6 RGB Images at 12 FPS and triggered a horizontal FOV of 360 degrees. Among them, recorded keyframes are annotated at 2Hz considering data synchronization between cameras and LiDAR. The mounted 32-beam LiDAR radiates the scene at 20 FPS. Following the convention of nuScenes authority, we transform the previous 9 frames into the current point cloud and form the points keyframe at 2Hz. 

We report the results on the two main metrics: mean Average Precision (mAP) and nuScenes detection scores (NDS). Besides, the results of 10 detection categories are also presented for a detailed comparison.


\noindent\textbf{Implementation details.}
We implement our network in PyTorch \cite{paszke2019pytorch} using the open-sourced MMDetection3D \cite{mmdet3d2020}. For the image branch, we use Dual-Swin-Tiny \cite{liang2022cbnet} with FPN \cite{lin2017feature} as the feature extractor and initialize it from the instance segmentation model Cascade Mask R-CNN \cite{cai2019cascade} pretrained on COCO \cite{lin2014microsoft} and then nuImage \cite{caesar2020nuscenes}, which is previously adopted by BEVFusion \cite{liangbevfusion}. For the LiDAR branch, VoxelNet \cite{zhou2018voxelnet} is chosen as the backbone. As for the LGVT module, we set the height of the pillar to four and the number of layers to three, respectively. For temporal fusion, we use a total of five frames as inputs. We set the image size to 448 × 800 and the voxel size to (0.075m, 0.075m, 0.2m) following \cite{liangbevfusion, bai2022transfusion, zhou2018voxelnet, lang2019pointpillars}. Our training consists of two stages: i) We first train a LiDAR-only detector for 20 epochs; ii) We then apply joint training to BEVFusion4D for another 6 epochs. During the second training stage, we freeze the weights of the image branch which is a common training strategy in previous methods \cite{bai2022transfusion, yang2022deepinteraction}. As for data augmentation, BEV-space data augmentation is adopted following \cite{liangbevfusion, bai2022transfusion}. Optimization is carried out using AdamW \cite{loshchilov2017decoupled} with learning rate of $2e^{-4}$ and weight decay of 10$^{-2}$. We do not use Test-Time Augmentation (TTA) or multi-model ensemble during inference.

\subsection{Comparisons with Previous Results}
Table~\ref{tab:nuscenes_val} and \ref{tab:nuscenes_test} list our results on par with previous competing methods in validation and test datasets, respectively. Our model without temporal fusion (\name{}-S) outperforms previous methods and achieves the detection scores of (70.9\% mAP, 72.9\% NDS) in the validation set and (71.9\% mAP, 73.7\% NDS) in the test set. Moreover, our method surpasses the previous state-of-the-art~\cite{yang2022deepinteraction} in most of the detection categories (9/10 of the validation and 7/10 of the test set). This overall performance gain could be attributed to our proposed LGVT module, which emphasizes the spatial prior of LiDAR on generating efficient camera BEV. We also note that our method improves the detection of certain categories (e.g. Truck) by a large margin. Since objects like trucks usually span multiple camera frames but are instead recorded in LiDAR as a whole, the prior of LiDAR provides a more complete spatial description of this type of object.

We also compare our spatiotemporal framework \name{} with previous state-of-the-art~\cite{zeng2022lift}. As indicated in Table~\ref{tab:nuscenes_test}, \name{} obtains an obvious performance gain by achieving 72.0\% mAP and 73.5\% NDS in the validation set and bringing a consistent improvement in 9/10 categories. In the test set, \name{} achieves the state-of-the-art performance of 73.3\% mAP and 74.7\% NDS, and rank 1st in mAP over most categories.

\subsection{Ablation Studies}
\textbf{The performance and generalization of the proposed LGVT.}
To understand the effect of our proposed LGVT module, we select a commonly used view transformation module LSS in ~\cite{liangbevfusion, liu2022bevfusion} as comparison and conduct studies on three detection heads, PointPillars~\cite{lang2019pointpillars}, CenterPoint~\cite{yin2021center} and TransFusion-L~\cite{bai2022transfusion}. As shown in Table~\ref{table:abl_view_transformer}, our result performs a consistent improvement in all settings and improves the result of LSS by 7.46\% mAP and 4.14\% NDS in PointPillar, 2.84\% mAP and 1.46\% NDS in CenterPoint and 1.65\% mAP and 0.98\% NDS in TransFusion-L, respectively. 
This implies that our method is able to generalize well in common configurations and exhibits superiority over the previous LSS due to the explicit guidance of LiDAR.
\begin{table}[h]	
\centering
\caption{Performance and Generalization of our view transformer module against the previous approach on nuScenes val set. }
\vspace{0.2cm}
\label{table:abl_view_transformer}
\renewcommand\arraystretch{1}
\setlength{\tabcolsep}{5pt}
\footnotesize
\begin{adjustbox}{max width=0.45\textwidth}
\begin{tabular}{*{7}{c}}
\toprule
 \vcenterhead{View Transform Module} & \multicolumn{2}{c}{PointPillars} & \multicolumn{2}{c}{CenterPoint} & \multicolumn{2}{c}{TransFusion-L} \\
 \cmidrule(l{.5\tabcolsep}r{.5\tabcolsep}){2-3}
 \cmidrule(l{.5\tabcolsep}r{.5\tabcolsep}){4-5}
 \cmidrule(l{.5\tabcolsep}r{.5\tabcolsep}){6-7}
  & mAP$\uparrow$ & NDS$\uparrow$ & mAP$\uparrow$ & NDS$\uparrow$ & mAP$\uparrow$ & NDS$\uparrow$\\
\midrule
LSS & 44.55  & 55.07 & 60.60 & 66.91 & 69.29 & 71.95 \\
\midrule
LGVT & 52.01 & 59.21 & 63.44 & 68.37 & 70.94 & 72.93 \\
\bottomrule
\end{tabular}
\end{adjustbox}
\normalsize
\end{table}

\begin{figure}[htbp]
\centering
\includegraphics[scale=0.45]{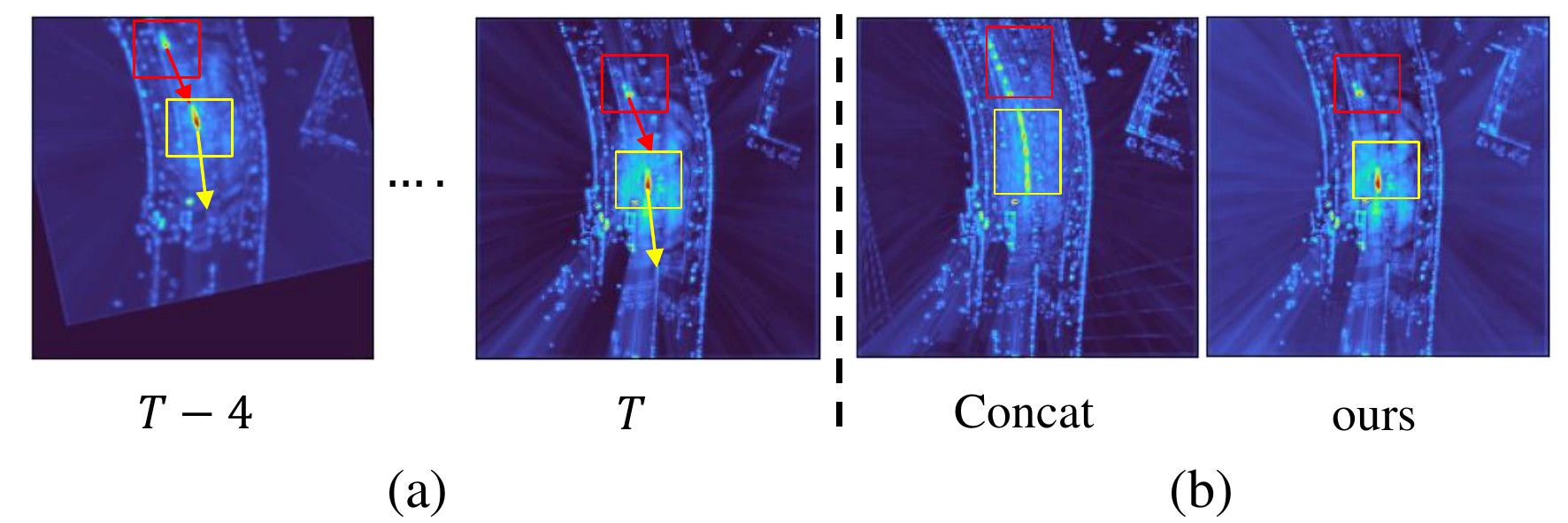}
\caption{(a) shows aligned fused BEV features $\Tilde{F}^{T-4}$ and $\Tilde{F}^T$. The red and yellow squares represent two moving objects, respectively. The arrow indicates the direction of the object's movement. (b) shows the result of different ways of temporal fusion, simply concatenating {$\Tilde{F}^{T-4},...,\Tilde{F}^T$} will cause motion smear, while our method can alleviate this effect.}
\label{Fig.motion_smear} 
\end{figure}

\textbf{Results of the proposed TDA module.}
We illustrate the result of our temporal fusion approach in Table~\ref{table:abl_tempo}. Our baseline without temporal fusion (row one) receives a score of 70.9\% mAP and 72.9\% NDS, respectively. Gradually increasing the length of the input frames leads to a gain up to 1.0\% mAP and 0.6\% NDS, which implies that our method is capable of aggregating historical information over a long time span. We also conduct the experiment using a naive fusion approach in which the temporal information is concatenated and passed through a convolution module. We observe a performance drop compared to our method. This corresponds to our findings in Fig.~\ref{Fig.motion_smear} that the deformable attention mechanism is beneficial to tackle the problem of motion smear in moving objects and yields better predictions in this case.
\begin{table}[h]	
\centering
\caption{The study of our temporal fusion module providing different lengths of frames on nuScenes val set. }
\vspace{0.2cm}
\label{table:abl_tempo}
\renewcommand\arraystretch{1}
\setlength{\tabcolsep}{5pt}
\footnotesize
\begin{tabular}{l|cc|cc}
\hline
& \multicolumn{2}{c|}{Naive} & \multicolumn{2}{c}{Ours}\\
\hline
 Length & mAP$\uparrow$ & NDS$\uparrow$   & mAP$\uparrow$      & NDS$\uparrow$  \\  
 \hline
  T = 1& 70.94 & 72.93   &  70.94    & 72.93  \\  
 \hline
  T = 3& 71.34 &  72.93  &   71.38   & 73.20  \\
 \hline
  T = 5& 71.77 &  73.35  &  71.95    & 73.50  \\  
 \hline
 
\end{tabular}
\end{table}

\textbf{Query strategy in LGVT.}
In Table ~\ref{table:query_query}, we compare various camera query strategies in LGVT to validate the necessity of our cross-modality query interaction. We first study the effect of query generation strategies given a single modality input ((a) - (c) in Table.~\ref{table:query_query}). (a) denotes only using the randomly initialized camera BEV as the query, which is used in~\cite{li2022bevformer}. (b) represents only using our initialized camera BEV as query, which is indicated in Eq.~\ref{eq.project}-\ref{eq.sample_selection} . In (c), the query is simply initialized by the LiDAR BEV. We observe that only using LiDAR BEV to initialize camera query (c) brings a performance gain over the camera-BEV-based approach. This could be attributed to the effective spatial information hidden in the LiDAR BEV, which delivers a more reliable reference to extract image features of relevant targets. Then we explore the compatibility with the LiDAR BEV in query generation strategies ((d) - (e) in Table ~\ref{table:query_query}). By enforcing LiDAR BEV into the camera query, we find that our method (e) results in a rise of performance compared with the random one (d). It indicates that the LiDAR BEV is better compatible with our sampled camera BEV since they are both input-dependent.


\begin{table}[h]	
\centering
\caption{The comparison between various camera query generation strategies in our proposed LGVT module. Results are evaluated on nuScenes val set.}
\vspace{0.2cm}
\label{table:query_query}
\renewcommand\arraystretch{1}
\setlength{\tabcolsep}{5pt}
\footnotesize
\begin{adjustbox}{max width=0.45\textwidth}
\begin{tabular}{*{6}{c}}
\toprule
  & \vcenterhead{LiDAR BEV} & \multicolumn{2}{c}{Camera BEV} &  \vcenterhead{mAP$\uparrow$} &  \vcenterhead{NDS$\uparrow$}  \\
   \cmidrule(l{.5\tabcolsep}r{.5\tabcolsep}){3-4}
    & & random & sampled & & \\
    \midrule
(a) &  & \checkmark &  & 70.27 & 72.54\\
(b) &  &  & \checkmark  & 69.76 & 72.47\\
(c) & \checkmark &  &   & 70.62 & 72.71 \\
(d) & \checkmark & \checkmark &   & 70.50 & 72.66\\
(e) & \checkmark  &  & \checkmark  &   70.94& 72.93\\
\bottomrule
\end{tabular}
\end{adjustbox}
\normalsize
\end{table}

\textbf{Model Capacity and Efficiency.} 
We compare our framework with the previous leading BEV fusion method~\cite{liangbevfusion} in terms of four criteria shown in Table~\ref{table:abl_fusion}. Without the temporal fusion module, our \name{}-S outperforms BEVFusion by a margin of 0.6\% mAP and 0.4\% NDS thanks to our LGVT module, while reducing the model size by around 21\% and achieves a FPS of 2.2. Our \name{} further exploits the cross-modality feature interaction in the spatiotemporal domain via our proposed temporal alignment module TDA, thus bringing a continuing improvement of 1.4\% mAP and 1\% NDS without introducing a noticeable memory cost and computation load.
\begin{table}[h]	
\centering
\caption{Comparison with previous leading BEV fusion method on nuScenes test set. BEVFusion4D-S represents our approach without temporal fusion. The FPS is tested on a single NVIDIA Tesla V100 GPU.}
\vspace{0.2cm}
\label{table:abl_fusion}
\renewcommand\arraystretch{1}
\setlength{\tabcolsep}{5pt}
\footnotesize
\begin{adjustbox}{max width=0.45\textwidth}
\begin{tabular}{lcccc}
\toprule
Method & mAP$\uparrow$ & NDS$\uparrow$ & Params & FPS$\uparrow$ \\
\midrule
BEVFusion~\cite{liangbevfusion} & 71.3 & 73.3 & 90.3M & 0.6 \\
\midrule
BEVFusion4D-S & 71.9 & 73.7 & 71.7M & 2.2 \\
BEVFusion4D & 73.3 & 74.7 & 72.0M & 2.0 \\
\bottomrule
\end{tabular}
\end{adjustbox}
\normalsize
\end{table}

\graphicspath{{\subfix{../Images/}}}

\section{Conclusion}
In this paper, we present a novel 3D object detection spatiotemporal framework named BEVFusion4D. In the spatial fusion stage, we design a LiDAR-Guided View Transformer (LGVT) to help obtain a better image representation in BEV space via the guidance of LiDAR prior. In the temporal fusion stage, we propose a Temporal Deformable Alignment (TDA) module to recurrently aggregate multiple frames with negligible additional memory cost and computation budget. Extensive experiments demonstrate the effectiveness of our two modules and our method finally achieves state-of-the-art performances on the nuScenes dataset. We hope that BEVFusion4D can serve as a powerful baseline to inspire future research on multi-sensor fusion.

{\small
\bibliographystyle{ieee_fullname}
\bibliography{egbib}
}

\setcounter{page}{1}
\clearpage
{
\section*{Supplementary Material}
\renewcommand\thesection{\Alph{section}}
\setcounter{section}{0}

The supplementary document is organized as follows:
\begin{itemize}
    \vspace*{-2mm}
    \item[-] Result details and analysis of the proposed TDA module for temporal feature aggregation.
    \vspace*{-2mm}
    \item[-] More ablation study results of the proposed LGVT module.
    \vspace*{-2mm}
    \item[-] Visualization of model's prediction results in both figure and video.
\end{itemize}

\section{Result details and analysis of TDA module}
\begin{table}[h]	
\centering
\caption{The statistic of 10 detection categories wrt. the speed ranges. Results are obtained in the nuScenes validation set.}
\vspace{0.2cm}
\label{table:speed_analysis}
\renewcommand\arraystretch{1}
\setlength{\tabcolsep}{5pt}
\footnotesize
\begin{adjustbox}{max width=0.5\textwidth}
\begin{tabular}{lccc}
\toprule
Class & 0 - 10 km/h & 10 - 30 km/h & $>$30 km/h \\
\midrule
Car & 41414 & 6811 & 4675\\
\midrule
Truck & 7745 & 1101 & 410\\\midrule
Bus & 1261 & 711 & 90\\\midrule
Trailer & 2180 & 128 & 23\\\midrule
C.V. & 1507 & 4 & 1\\\midrule
Pedestrian & 23402 & 48 & 0\\\midrule
Motorcycle &  1394 & 332 & 128\\\midrule
Bicycle & 1623 & 197 & 2\\\midrule
Traffic cone & 10523 & 1 & 0\\\midrule
Barrier & 15888 & 5 & 0\\\bottomrule

\end{tabular}
\end{adjustbox}
\normalsize
\end{table}
\begin{table}[h]	
\centering
\caption{Comparison of the model's performance (Naive/TDA) under different speed ranges. TDA achieves an overall gain in 10-30 km/h and $>$30 km/h cases in terms of average precision (AP). Results are evaluated on nuScenes validation set.}
\vspace{0.2cm}
\label{table:speed_result}
\renewcommand\arraystretch{1}
\setlength{\tabcolsep}{5pt}
\footnotesize
\begin{adjustbox}{max width=0.5\textwidth}
\begin{tabular}{*{4}{c}}
\toprule
  \vcenterhead{class} & \multicolumn{3}{c}{AP$\uparrow$ (Naive/TDA)} \\
   \cmidrule(l{.5\tabcolsep}r{.5\tabcolsep}){2-4} 
    & 0-10 km/h& 10-30 km/h& $>$30 km/h \\\midrule
    Car & 0.906/0.906	& 0.902/0.907 & 0.907/0.907\\\midrule
    Truck & 0.703/0.692	&0.753/0.759&	0.730/0.747\\\midrule
    Bus & 0.829/0.827&	0.774/0.808&	0.691/0.718\\\midrule
    Motorcycle & 0.813/0.804&	0.817/0.857&	0.842/0.855\\\midrule
    Average& \textbf{0.813}/0.807 & 0.812/\textbf{0.833} & 0.793/\textbf{0.807} \\

\bottomrule
\end{tabular}
\end{adjustbox}
\normalsize
\end{table}

\begin{table}[h]	
\centering
\caption{Comparison of the model's performance (Naive/TDA) under different speed ranges. TDA achieves an overall gain in 10-30 km/h and $>$30 km/h cases in terms of average velocity error (AVE). Results are evaluated on nuScenes validation set.}
\vspace{0.2cm}
\label{table:speed_result}
\renewcommand\arraystretch{1}
\setlength{\tabcolsep}{5pt}
\footnotesize
\begin{adjustbox}{max width=0.5\textwidth}
\begin{tabular}{*{4}{c}}
\toprule
  \vcenterhead{class} & \multicolumn{3}{c}{AVE$\downarrow$ (Naive/TDA)}  \\
   \cmidrule(l{.5\tabcolsep}r{.5\tabcolsep}){2-4}
    & 0-10 km/h& 10-30 km/h& $>$30 km/h\\\midrule
    Car & 0.105/0.106 &	0.681/0.659 & 1.015/0.963\\\midrule
    Truck	&0.103/0.111&	0.825/0.802&	1.098/1.124\\\midrule
    Bus &	0.148/0.151&	0.798/0.766&	1.702/1.660\\\midrule
    Motorcycle &	0.109/0.108&	0.846/0.762& 2.372/2.234\\\midrule
    Average & \textbf{0.116}/0.119 & 0.788/\textbf{0.747} & 1.547/\textbf{1.496}\\

\bottomrule
\end{tabular}
\end{adjustbox}
\normalsize
\end{table}

We present an extra result comparison between our TDA module and the naive approach mentioned in section 4.3 of the paper. To understand how TDA helps remedy the motion smear effect caused by the moving object between frames, we conduct experiments on the validation set of the nuScenes dataset and divide the ground truth box lists into three speed ranges: 1) 0-10 km/h, 2) 10 - 30 km/h, and 3) $>$ 30 km/h, which represent ground-truth objects at low, mid, and high speed, respectively. The statistic of 10 detection categories is shown in Table~\ref{table:speed_analysis}. It could be observed that objects like cars and trucks move at various speeds, while the speed of pedestrians mainly falls in the 0-10 km/h. To reduce bias in the calculation, we keep four main categories (car, truck, bus, and motorcycle) that span all speed ranges with a certain number and conduct the model performance study on them.  Following nuScenes protocol, we provide the result based on the evaluation of average precision (AP) and average velocity error (AVE) metrics. As shown in Table~\ref{table:speed_result}, incorporating the TDA module for temporal feature aggregation leads to a performance improvement in the mid/high-speed categories. Since the increase in speed will cause a more prominent motion smear effect, this poses several problems to the system that could not be alleviated by the naive approach. Compared to that, our attention-based TDA module learns from a sparse deformable feature sampling, which may benefit the temporal feature aggregation of moving objects due to the advantage of the deformable attention mechanism. Therefore, TDA is able to infer the actual position and speed of objects given better-aligned temporal features. We also note that our TDA module achieves a similar result as the naive approach in the low-speed case (0-10 km/h). Moreover, since most of the objects in nuScenes dataset fall into a low-speed category, the benefit of applying TDA is comparably marginal in terms of overall mAP and NDS metrics. However, our study reveals an essential fact that is seldom discussed in the previous paper and indicates that a dedicated TDA module is more favorable in the practical setting where the speed of moving objects is more diverse.

\section{More ablation study results of LGVT module}
\begin{table}[h]	
\centering
\caption{Ablations on the number of layers in LGVT module (section 3.2). Results are performed on nuScenes validation set.}
\vspace{0.2cm}
\label{table:layers_study}
\renewcommand\arraystretch{1}
\setlength{\tabcolsep}{5pt}
\footnotesize
\begin{adjustbox}{max width=0.5\textwidth}
\begin{tabular}{lccccccc}
\toprule
Layers & mAP$\uparrow$ & NDS$\uparrow$ & mATE$\downarrow$ & mASE$\downarrow$ & mAOE$\downarrow$ 
& mAVE$\downarrow$ & mAAE$\downarrow$\\
\midrule
1	& 70.51	& 72.57	& 0.266	& 0.251	& 0.316	& 0.255	& 0.181\\\midrule
2	& 70.61	& 72.78	& 0.263	& 0.248	& 0.302	& 0.251	& 0.189\\\midrule
3	& \textbf{70.94}	& \textbf{72.93}	& 0.264	& 0.249	& 0.304 &	0.254 &	0.183\\
\bottomrule

\end{tabular}
\end{adjustbox}
\normalsize
\end{table}
\begin{table}[h]	
\centering
\caption{Ablations on the number of pre-defined BEV heights in LGVT module (section 3.2). Results are performed on nuScenes validation set.}
\vspace{0.2cm}
\label{table:height_study}
\renewcommand\arraystretch{1}
\setlength{\tabcolsep}{5pt}
\footnotesize
\begin{adjustbox}{max width=0.5\textwidth}
\begin{tabular}{lccccccc}
\toprule
Heights & mAP$\uparrow$ & NDS$\uparrow$ & mATE$\downarrow$ & mASE$\downarrow$ & mAOE$\downarrow$ 
& mAVE$\downarrow$ & mAAE$\downarrow$\\
\midrule
1	& 70.45	& 72.74	& 0.267	& 0.250	& 0.293	& 0.258	& 0.181\\\midrule
2	& 70.33	& 72.65	& 0.263	& 0.249	& 0.305	& 0.256& 	0.184\\\midrule
4	& \textbf{70.94}	& \textbf{72.93}	& 0.264	& 0.249	& 0.304	& 0.254	& 0.183\\\midrule
8	& 70.48	& 72.72 & 0.268	& 0.249	& 0.295 &	0.257 &	0.183\\
\bottomrule

\end{tabular}
\end{adjustbox}
\normalsize
\end{table}
We review the LGVT module with more details and related studies. 

\textbf{Ablations on the number of layers in LGVT.} In Table~\ref{table:layers_study}, we study the effect of the number of layers on the system's performance. It could be obtained that LGVT with one layer achieves a baseline result of 70.51\% mAP and 72.57\% NDS. Gradually adding the number of layers leads to a rise up to 70.94\% mAP (+0.43) and 72.93\% NDS (+0.36). This implies that repeatedly augmenting camera BEV with LiDAR prior via our LGVT module facilitates extracting more corresponding image features and thus brings an observable gain.

\textbf{Ablations on the number of heights in LGVT.} 
We also investigate the choice of pre-defined BEV heights in LGVT. Table~\ref{table:height_study} represents the experimental results, where the number of heights of the pre-defined BEV grid (section 3.2) is set as 1,2,4, and 8, respectively. Specifically, we assign $N$ evenly spaced heights on top of the BEV grid and project each reference point into the image plane following a similar principle of ~\cite{li2022bevformer}. The comparison between different height numbers indicates that applying 4 heights to each BEV query gives the best performance, which is adopted as our final choice in the paper. We conjugate that a proper number of height set up helps LGVT query relevant image features efficiently. 

\section{Visualization}

\textbf{Visual comparisons between our BEVFusion4D and BEVFusion~\cite{liangbevfusion}.}
We compare the visual results of BEVFusion4D and the previous work BEVFusion~\cite{liangbevfusion} in Fig.~\ref{Fig.scence_cmp.pdf}. BEVFusion4D is able to detect objects at various distances thanks to our LGVT and TDA modules for spatial and temporal fusion, respectively.


\begin{figure*}[htbp]
\centering
\includegraphics[width=\textwidth]{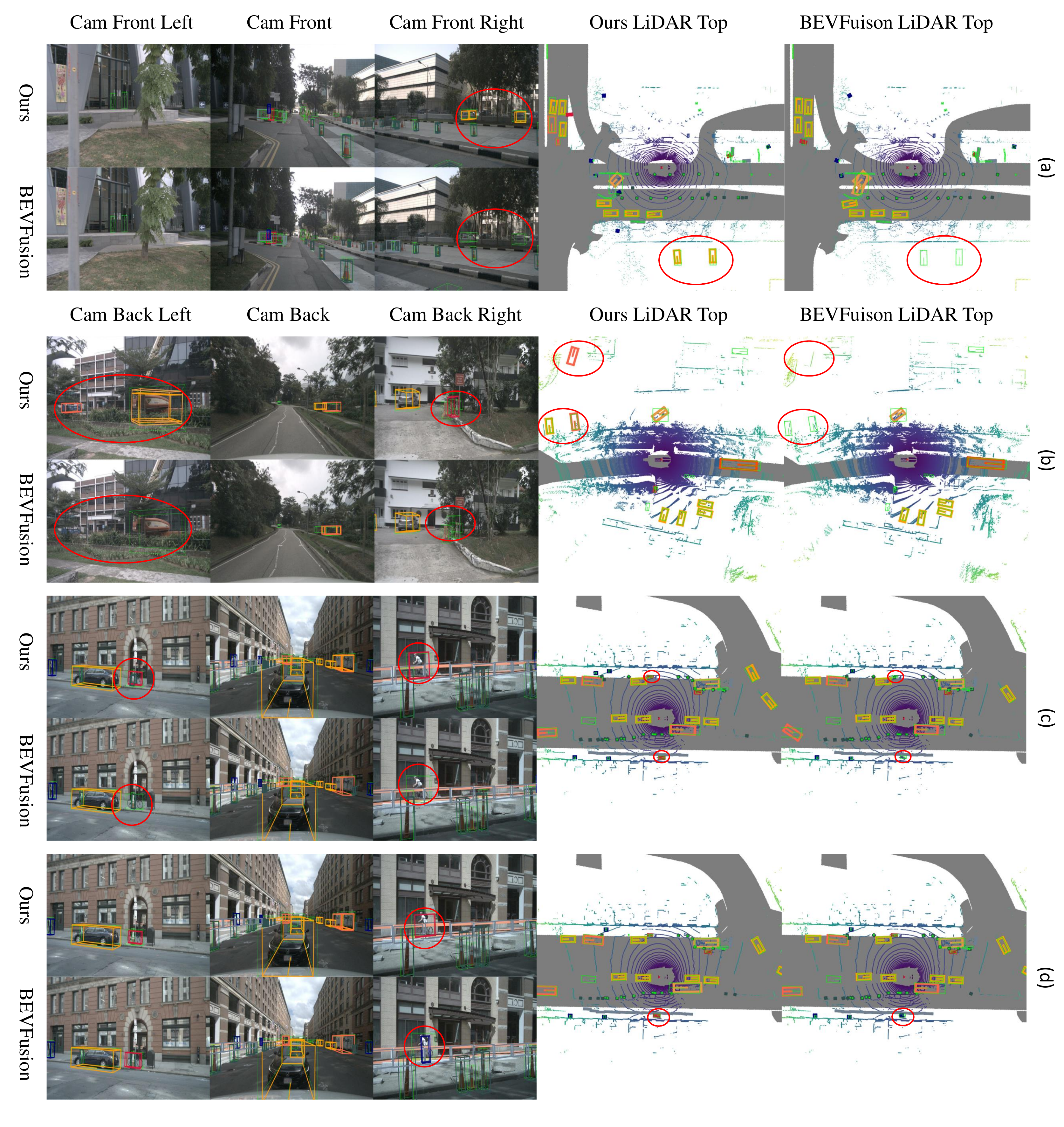}
\caption{Visual comparisons between BEVFusion4D and BEVFusion~\cite{liangbevfusion} on nuScenes validation set. All green rectangular boxes in the figures represent ground truth, while the other colored rectangular boxes represent the predicted results. The red circled areas in (a) demonstrate the model's detection ability for distant targets; the red circled areas in (b) demonstrate the model's detection ability for occluded targets; (c) and (d) represent two sequential frames of the same scene, and the red circled areas reflect the improved results achieved by our model after incorporating temporal information.}
\label{Fig.scence_cmp.pdf} 
\end{figure*}

}

\end{document}